\documentclass[letterpaper, 10 pt, conference]{class/ieeeconf}  

\IEEEoverridecommandlockouts                              
\usepackage{siunitx}
\usepackage[table,x11names]{xcolor}
\usepackage{graphicx}

\usepackage{amsmath}
\usepackage{pifont}
\usepackage{tcolorbox}
\usepackage{enumerate}
\let\labelindent\relax
\usepackage[shortlabels]{enumitem}\usepackage{amssymb}
\usepackage{latexsym}
\usepackage{url}
\usepackage{cite}
\usepackage{relsize}
\usepackage{multirow}
\usepackage{afterpage}
\usepackage{ifthen}
\usepackage{graphicx}
\usepackage{algpseudocode,algorithm,algorithmicx}
\usepackage{subfigure}
\usepackage{flushend}
\usepackage{epstopdf}
\usepackage{stackengine}
\usepackage{textcomp} 
\usepackage{pifont}
\usepackage{threeparttable,booktabs}
\usepackage{gensymb}
\usepackage{booktabs}
\usepackage{makecell}
\usepackage{hhline}
\usepackage{courier}
\usepackage{lipsum}
\usepackage{diagbox}
\usepackage{xcolor} 
\usepackage{xspace}
\usepackage{bbm}

\usepackage{xspace}

\newcommand{\nbf}[1]{{\noindent \textbf{#1}}}

\makeatletter
\let\NAT@parse\undefined
\makeatother
\usepackage[bookmarks=false, linkcolor=blue, urlcolor=blue, citecolor=blue]{hyperref} 
 \graphicspath{{./figures/}}

\hypersetup{
    colorlinks=true,
    linkcolor=red,
    filecolor=magenta,      
    urlcolor=blue,
    pdfstartview={FitH},
    citecolor =blue
    }

\title{\LARGE \bf
eVGGT: An Efficient Geometry-Aware Vision Encoder for \\ Visuomotor Policies}

\author{
An Dinh Vuong$^{1}$, Minh Nhat Vu$^{2,3}$, Ian Reid$^{1}$
\thanks{$^{1}$Department of Computer Vision,
        Mohamed bin Zayed University of Artificial Intelligence, UAE
        {\tt\small an.vuong@mbzuai.ac.ae}
        }%
\thanks{$^2$ AIT Austrian Institute of Technology GmbH, Austria  
}
\thanks{$^3$
Center for AI Research (CAIR), VinUniversity, Vietnam
}
}

\begin{document}

\maketitle

\thispagestyle{empty}
\pagestyle{empty}



\begin{abstract}
Geometry-grounded vision models, such as VGGT, have emerged as robust visual encoders, providing essential geometric priors for robotic manipulation. However, the high computational cost of these models often leads to slow inference, limiting their practical applications in real-world robotics. This paper introduces \textbf{eVGGT}, a lightweight geometry-aware vision encoder distilled from the high-performing VGGT. Our findings demonstrate two primary advantages: \textit{i)} integrating eVGGT into imitation learning frameworks (including ACT and Diffusion Policy) yields up to a 6.3\% improvement in success rate over standard 2D encoders across bimanual and single-arm tasks in both simulation and real-world settings with variable viewpoints; \textit{ii)} eVGGT achieves a nearly $5\times$ speedup and a 63\% reduction in memory usage compared to state-of-the-art geometry-aware encoders while maintaining comparable task performance. These results suggest that eVGGT substantially alleviates the performance-latency bottleneck that has limited geometry-aware visuomotor policies in real-world deployment. Code and checkpoints are available at \href{https://evggt.github.io/}{https://evggt.github.io/}.



\end{abstract}


\section{INTRODUCTION}
Understanding 3D geometry is fundamental to robotic manipulation, as it enables policies to infer object shape, pose, and contact-relevant structure from sensory information~\cite{jia2025lift3d, patel2026defm, yang2026robo3r}. Prior work injects explicit 3D structure into visuomotor policies by encoding point clouds or other geometric primitives~\cite{guo2025glad, lu2024manigaussian}. However, these pipelines typically rely on calibrated camera setups or depth sensors, which are costly and may be unavailable in real-world conditions~\cite{liu2026twin}. Motivated by these limitations, recent literature has shifted toward vision-only foundation models that capture geometric representations from multi-view RGB observations~\cite{ge2025vggt, lin2025evo, qian2025gp3}.

Visual geometry feed-forward models~\cite{murai2025mast3r, wang2024dust3r, wang2025vggt, keetha2025mapanything} have advanced geometric perception for robotics~\cite{guo2025glad}. By lifting 2D visual inputs into 3D representations, these models help imitation learning (IL) overcome viewpoint ambiguity in standard visuomotor policies~\cite{abouzeid2025geoaware}, improving performance in complex manipulation tasks~\cite{patel2026defm}. Nevertheless, this geometric awareness comes at substantial computational cost~\cite{ge2025vggt}. Specifically, integrating VGGT~\cite{wang2025vggt} into visuomotor policies often results in a high-latency bottleneck~\cite{zhuo2025streaming}; indeed, our reimplementation achieves an inference rate of just 4.5\,Hz. This is consistent with prior findings~\cite{lin2025evo, ge2025vggt} and underscores the difficulty of real-time deployment. On the other hand, state-of-the-art manipulation policies explicitly constrain real-world execution to 20\,Hz to maintain stable operation \cite{wang2025one, nguyen2025graspmamba}. Consequently, roboticists face a critical dilemma: prioritize high-fidelity geometric reasoning at the expense of policy update frequency, or maintain real-time responsiveness by sacrificing 3D spatial understanding.

\begin{figure}
    \centering
    \includegraphics[width=\linewidth]{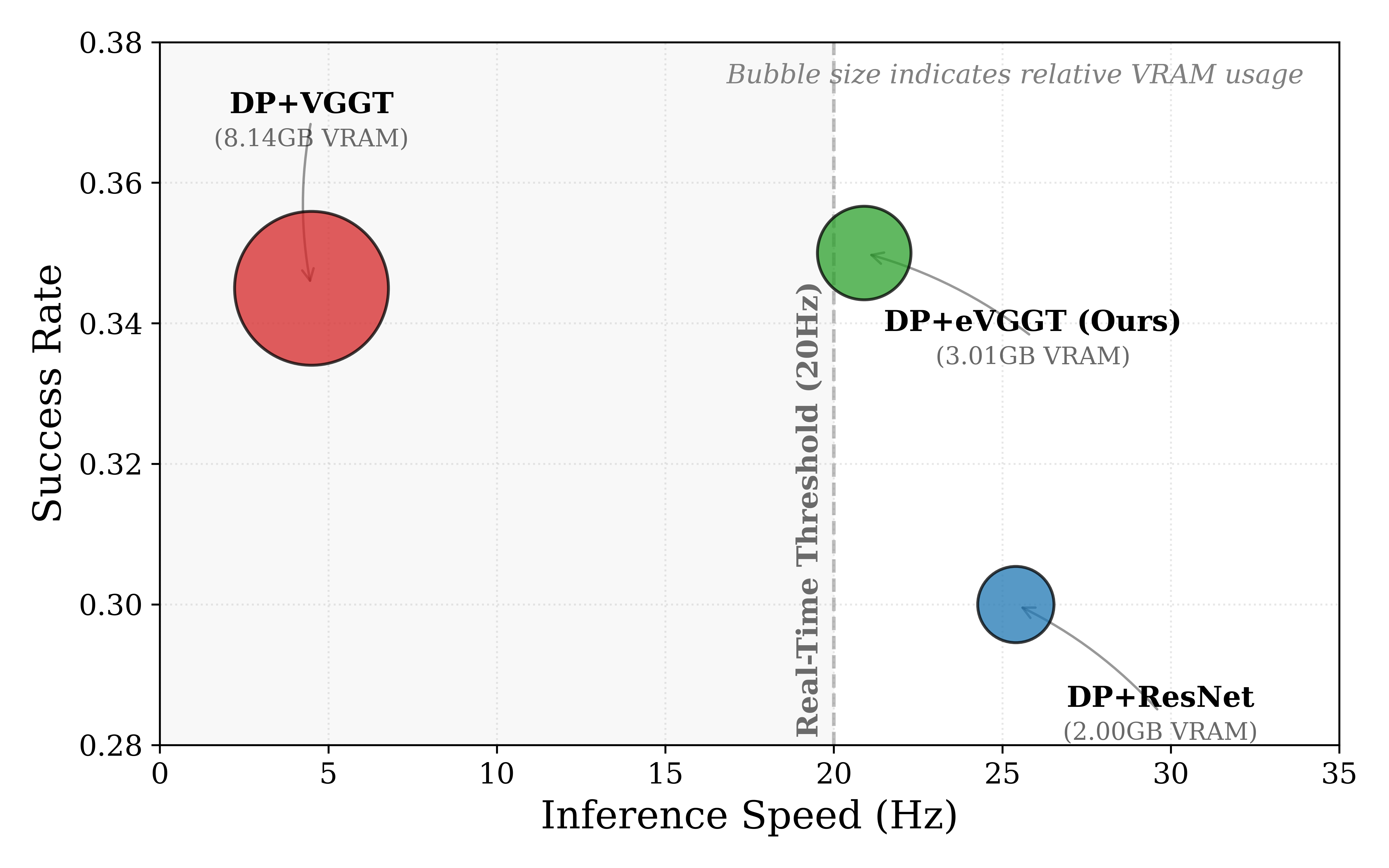}
    \caption{\textbf{Performance-efficiency trade-off of vision encoders.} We compare DP~\cite{chi2023diffusion}, the standard visuomotor policy, paired with different vision backbones. Success rate ($y$-axis) vs. inference throughput in\,Hz ($x$-axis); bubble area denotes VRAM (GB). The dashed line marks the 20\,Hz real-time threshold. DP+VGGT is robust but slow and memory-heavy (8.14\,GB, 4.5\,Hz); DP+eVGGT achieves the best trade-off with the highest success rate, enabling real-time deployment at lower memory (3.01 GB, 20.9 \,Hz).}
    \label{fig:teaser-evggt}
\end{figure}

Although recent efforts have explored faster visual geometry models~\cite{shen2025fastvggt, zhuo2025streaming, feng2025quantized}, these approaches primarily target conventional 3D reconstruction and remain too memory-intensive for real-time robotic policies. In this work, we address the latency-geometry trade-off by proposing eVGGT, a low-latency, geometry-aware vision encoder optimized for robotic manipulation. Specifically, we distill VGGT’s explicit geometric reconstructions, including camera poses, depth, and 3D point clouds, into a compact student model via knowledge distillation (KD), substantially reducing inference cost while preserving high-fidelity 3D reconstruction. As illustrated in Fig.~\ref{fig:teaser-evggt}, eVGGT delivers a nearly 5$\times$ speedup over VGGT, reaching 20.9\,Hz, without compromising the high manipulation success rate.

Integrating eVGGT into standard visuomotor frameworks, such as DP~\cite{chi2023diffusion} and ACT~\cite{zhao2023learning}, is straightforward. By replacing conventional 2D encoders, such as ResNet or ViT, with eVGGT, we provide policies with latent geometric features pivotal for complex manipulation~\cite{ge2025vggt}. We evaluate our approach across diverse simulation environments and real-world robots under varying view counts, spanning multi-view bimanual manipulation in RoboTwin~\cite{chen2025robotwin} and single-view single-arm manipulation in ManiSkill~\cite{taomaniskill3}. Across these settings, our framework achieves success rates comparable to state-of-the-art geometry-aware policies~\cite{abouzeid2025geoaware, qian2025gp3, lin2025evo}. Our contributions are threefold:

\begin{itemize}[leftmargin=*]
    \item We present eVGGT, a lightweight, geometry-aware encoder distilled from VGGT that retains robust geometric reasoning while achieving a nearly 5$\times$ speedup for visuomotor policies.
    \item We introduce a plug-and-play integration that injects eVGGT latent 3D features into standard visuomotor policies, such as DP and ACT, improving spatial reasoning without modifying the policy backbone.
    \item Across simulation and real-robot manipulation, our framework improves by up to $6.3\%$ over traditional 2D vision encoders, thanks to strong geometric priors from eVGGT.
\end{itemize}

\section{RELATED WORK}
\label{sec:related_work}

\nbf{Visual Representation for Imitation Learning.} Standard IL policies typically rely on 2D backbones like ResNet or ViT to encode visual observations~\cite{zhao2023learning, chi2023diffusion, liurdt}. While efficient, these 2D encoders often fail to capture spatial dynamics and suffer from viewpoint ambiguity~\cite{donattowards}. To address this, recent literature has incorporated explicit 3D representations, such as point clouds~\cite{qin2023dexpoint, ze20243d} and Gaussian splatting~\cite{lu2024manigaussian}, which directly encode scene geometry. However, these explicit methods introduce significant complexity and often require multi-stage pre-processing or external foundation models~\cite{ze20243d}, making them difficult to scale for high-frequency control~\cite{donattowards, xu2025flying}. This has motivated more direct and efficient geometric understanding, where 3D structure is extracted from sensory observations in a feed-forward manner, reducing reliance on heavy multi-stage pipelines~\cite{li2025geometry}.

\nbf{Geometry-Aware Visuomotor Policies.} Recent work has shifted towards leveraging latent features from high-performing feed-forward visual geometry networks, such as Evo-0~\cite{lin2025evo}, GP3~\cite{qian2025gp3}, and others~\cite{abouzeid2025geoaware, guo2025glad}. Many of these approaches utilize VGGT as a drop-in replacement for conventional 2D encoders within existing policy architectures; for instance, VGGT-DP~\cite{ge2025vggt} builds directly upon DP~\cite{chi2023diffusion}. Although these approaches are effective at capturing 3D spatial structure, their high latency~\cite{lin2025evo} limits their practical deployments in real-world robotic scenarios. We address this gap by introducing a lightweight geometry-aware network that facilitates efficient transfer to visuomotor policies.

\nbf{Imitation Learning for Robotic Manipulation.} IL is a promising approach for enabling generalizable manipulation behaviors using large-scale datasets~\cite{liurdt}. Given the difficulties of precisely determining environmental states~\cite{mandlekar2022matters}, 2D image-based IL methods such as ACT~\cite{zhao2023learning} and DP~\cite{chi2023diffusion} have emerged as effective and popular data-driven solutions in the field~\cite{taomaniskill3}. More recent IL approaches have shifted towards 3D-based representations to capture the structure of the environment better~\cite{chen2025robotwin, hoanggeometry}. By addressing efficiency in high-dimensional control, DP3~\cite{ze20243d} has proven to be among the most effective methods. Another prospective direction is the development of large-scale foundation models for multi-task generalist policies~\cite{etukuru2025robot}, such as RDT~\cite{liurdt}. However, these pretrained models~\cite{kim2026cosmos} often remain inefficient and require substantial fine-tuning to learn new tasks~\cite{di2024keypoint}.



\nbf{Efficient Visual Geometry Networks}. Prior work on accelerating visual geometry networks has primarily explored post-training compression\cite{feng2025quantized} and token pruning~\cite{shen2025fastvggt}. While these techniques reduce inference latency for offline 3D reconstruction, their number of network parameters often remains prohibitive for resource-constrained robotic platforms. KD is a well-established approach to model compression~\cite{mansourian2025comprehensive}. Although recent studies have used KD to transfer geometric reasoning from VGGT to downstream tasks such as video generation \cite{wu2025geometry}, our work focuses on leveraging KD to improve the computational efficiency of the VGGT architecture directly. To mitigate the risk of the student model overfitting to the teacher’s behavior \cite{he2022knowledge}, we employ an asymmetric data augmentation strategy: the teacher observes clean inputs. In contrast, the student learns to reconstruct the teacher’s geometric representations from distorted images, which improves robustness and generalization~\cite{wang2022makes}.

%

\section{METHOD}
\label{sec: method}
We present an efficient framework for integrating a visual geometry network into robotic manipulation. We introduce a knowledge-distilled variant of the VGGT that prioritizes computational efficiency while preserving critical geometric reasoning capabilities. The remainder of this section is organized as follows: first, we provide the preliminaries on the base VGGT architecture; next, we detail our KD framework; and finally, we describe how the resulting distilled encoder is integrated into visuomotor robotic policies.

\subsection{Preliminaries on VGGT}

\begin{figure}[!ht]
    \centering
    \includegraphics[width=\linewidth]{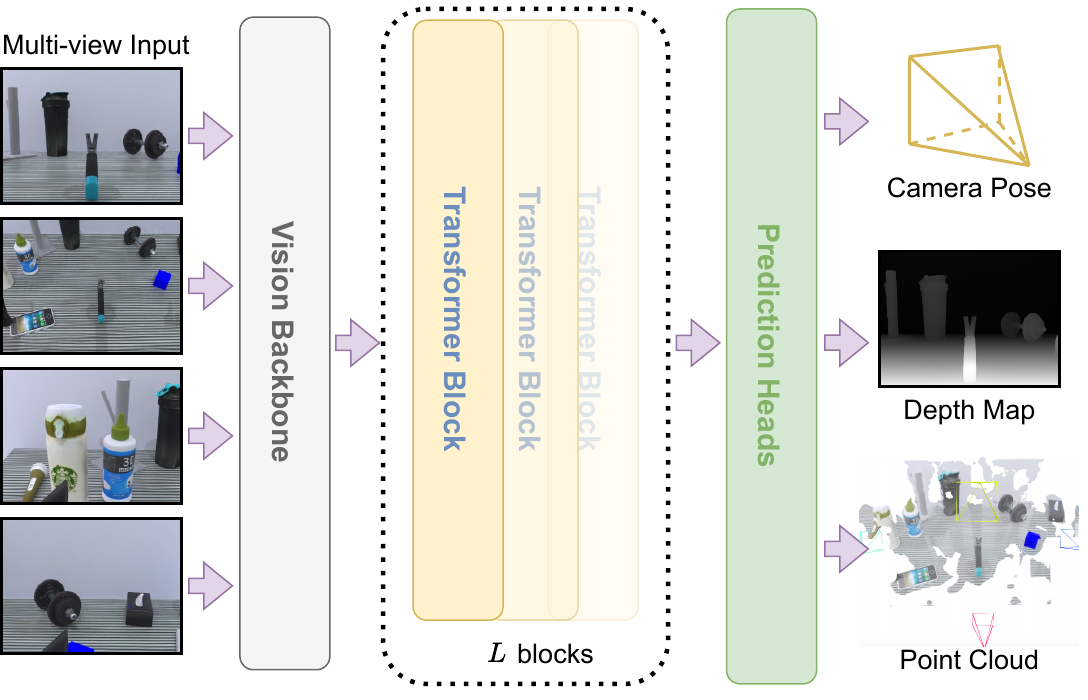}
    \caption{\textbf{VGGT architecture.} VGGT is a transformer-based vision encoder that produces geometry-aware features from multi-view input.}
    \label{fig: vggt-architecture}
\end{figure}

VGGT is a vision transformer trained on extensively large datasets and exhibits strong geometric reasoning capabilities compared to several other geometry-grounded networks~\cite{wang2024dust3r, murai2025mast3r, maggio2025vggt}. An overview of VGGT is depicted in Fig.~\ref{fig: vggt-architecture}.

Given a set of \(N\) input images \(\{I_1, I_2, \dots, I_N\}\), VGGT first tokenizes into patches of size \(P \times P\) using a fine-tuned DINOv2-ViT-L backbone~\cite{oquab2024dinov2}, producing frame token sequences \(\{\mathbf{\ell}_1, \mathbf{\ell}_2, \dots, \mathbf{\ell}_N\}\). To facilitate camera parameter prediction, special \emph{camera} tokens \(\mathbf{\ell}_i^{\text{c}}\) and \emph{register} tokens \(\mathbf{\ell}_i^{\text{r}}\) are added to each frame token sequence, allowing the model to distinguish the first frame from subsequent frames~\cite{darcetvision}. The final frame representation is denoted as \(\mathbf{z}_i = (\mathbf{\ell}_i, \mathbf{\ell}_i^{\text{c}}, \mathbf{\ell}_i^{\text{r}})\), and the collection of frame features is \(\mathcal{Z} = \{\mathbf{z}_1, \mathbf{z}_2, \dots, \mathbf{z}_N\}\).  

The feature set \(\mathcal{Z}\) is then fed into a transformer network comprising \(L\) transformer blocks. Each block employs \emph{Alternating-Attention}, which alternates between global attention over the entire sequence \(\mathcal{Z}\) and frame-wise self-attention applied to the individual frame token sequences \(\mathbf{z}_i, \forall i \in [1, N]\). The resulting output tokens \(\hat{\mathbf{z}}_i\) for each frame can then be utilized either by \textit{i)} a camera head, implemented as a transformer for estimating camera poses, or \textit{ii)} Dense Prediction Transformer (DPT) heads~\cite{ranftl2021vision}, which generate geometric outputs including depth maps and point clouds.

The main computational bottleneck of VGGT derives from its transformer module, comprising \(L=24\) blocks, which limits its practicality for robotic tasks. Moreover, VGGT is initially designed for high-resolution images, whereas most robotic applications use lower-resolution images to maintain efficiency and scalability~\cite{puighabitat}. Following the scaling practice~\cite{yao2024hiri}, we propose a compressed VGGT architecture that suits the lower-resolution images typical in these robot settings, described in Sec.~\ref{sec: method-eVGGT}.


\subsection{Efficient VGGT via Knowledge Distillation}\label{sec: method-eVGGT}
We present an efficient VGGT model trained via a KD scheme. First, we redesign a compact VGGT architecture, and then describe the training procedure used to obtain our lightweight model.

\nbf{Redesigned Architecture.} To improve the efficiency of VGGT, we reduce the number of transformer blocks from \(L=24\) to \(L=4\). The choice of $4$ blocks is motivated by VGGT’s design, where the DPT heads extract and fuse features from four intermediate transformer blocks to predict dense depth and point maps. To maintain compatibility with the teacher VGGT model, we preserve the structure of the prediction heads, making $4$ the minimal number of blocks that allows supervision of the student’s intermediate representations. In addition, since robotic environments typically operate on low-resolution images, we follow the scaling rules of ViT models~\cite{yao2024hiri} and replace the DINOv2-ViT-L backbone with the smaller DINOv2-ViT-S. These modifications result in a lightweight VGGT that preserves geometric reasoning while substantially lowering computational cost. We refer to this proposed architecture as \textbf{eVGGT} (\textbf{e}fficient \textbf{VGGT}), which serves as the student model trained under a pretrained VGGT teacher model. Importantly, the student relies only on the teacher’s outputs, requiring no ground-truth labels, making eVGGT applicable to diverse robotic environments.

\nbf{Training Knowledge Distillation.} The outputs of the VGGT model for each image $I_i$ include: \textit{i)} camera parameters \(\theta_i\), \textit{ii)} depth map \(\mathbf{D}_i\), and \textit{iii)} point cloud \(\mathbf{P}_i\). Denote \(\ast^{\text{std}}\) and \(\ast^{\text{tch}}\) as the outputs of the student and teacher models, respectively. Following~\cite{he2022knowledge}, the KD loss is defined as:

\begin{equation}
\label{eq: distillation_loss}
    \mathcal{L}_{\text{distill}} = \sum_{i=1}^N \Big( \|\theta_i^{\text{std}} - \theta_i^{\text{tch}}\|^2 + \|\mathbf{D}_i^{\text{std}} - \mathbf{D}_i^{\text{tch}}\|^2 + \|\mathbf{P}_i^{\text{std}} - \mathbf{P}_i^{\text{tch}}\|^2 \Big).
\end{equation}

Empirically, we observe that the depth maps predicted by the student model trained with Eq.~\eqref{eq: distillation_loss} exhibit noticeable noise and lack smoothness. This problem is mainly due to the scale difference between the student backbone (DINOv2-ViT-S) and the teacher backbone (DINOv2-ViT-L)~\cite{tian2024visual}. To address this, we incorporate a gradient loss following~\cite{wang2025vggt}:

\begin{equation}\label{eq: grad_loss}
    \mathcal{L}_{\text{grad}} = 
    \sum_{i=1}^N \Big(
        \|\nabla_x \mathbf{D}_i^{\text{std}} - \nabla_x \mathbf{D}_i^{\text{tch}}\|_1 +
        \|\nabla_y \mathbf{D}_i^{\text{std}} - \nabla_y \mathbf{D}_i^{\text{tch}}\|_1
    \Big),
\end{equation}
where \(\nabla_x\) and \(\nabla_y\) are the gradients along the $x$ and $y$ axes of the depth map. The final training loss is 
\(\mathcal{L} = \mathcal{L}_{\text{distill}} + \mathcal{L}_{\text{grad}}\).

\nbf{Data Augmentation.} 
Since data augmentation can improve the generalization of the student model~\cite{wang2022makes}, we augment only the student’s input images, keeping the teacher’s inputs unchanged for stable supervision. Following~\cite{wang2025vggt}, we adopt three augmentations: \textit{i)} color jitter (brightness, contrast, saturation, hue), \textit{ii)} random grayscale conversion with probability $0.3$, and \textit{iii)} Gaussian blur (a $3\times3$ kernel, $\sigma\in[0.1,2.0]$). These commonly used augmentations have proven effective for 3D reconstruction~\cite{xian2024towards}. As shown in Sec.~\ref{sec: experiment}, this simple strategy can significantly improve 3D reconstruction quality.

\subsection{Integration into Robotic Policies}\label{sec: policy}
To leverage the geometric reasoning capabilities of eVGGT for robotic policy learning, we replace the standard vision encoders in existing IL with a pretrained, frozen eVGGT, as described in Sec.~\ref{sec: method-eVGGT}. This frozen design choice enables direct use of 3D spatial representations without the need for additional task-specific fine-tuning, which is more suitably handled by the policy heads~\cite{jiangrobots}. We remove eVGGT’s prediction heads and extract only the latent geometric features from the transformer blocks (Fig.~\ref{fig: vggt-architecture}), substantially reducing computational cost. Adaptive MLPs are then applied to project these latent features to match the input dimensionality of the policy heads.

\begin{figure}[!ht]
    \centering
    \includegraphics[width=\linewidth,trim=0.5cm 0 0.5cm 0,]{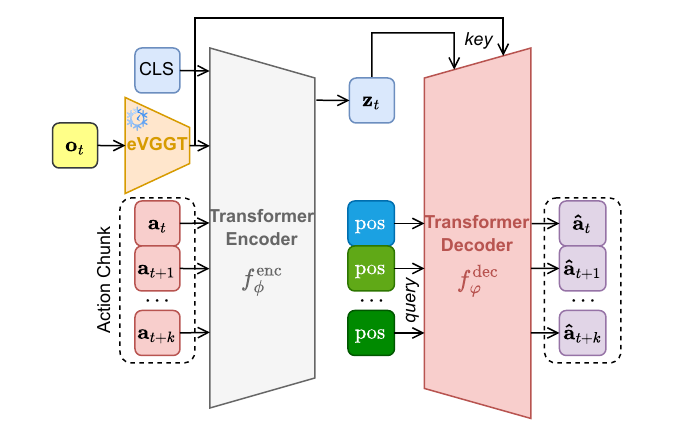}
    \caption{\textbf{Integration of eVGGT into ACT.} eVGGT provides geometry-aware visual features for both the encoder and decoder of ACT.}
    \label{fig: act-policy}
    \vspace{-1ex}
\end{figure}

\begin{figure}[!ht]
    \centering
    \includegraphics[width=\linewidth,trim=0.5cm 0 0 0,clip]{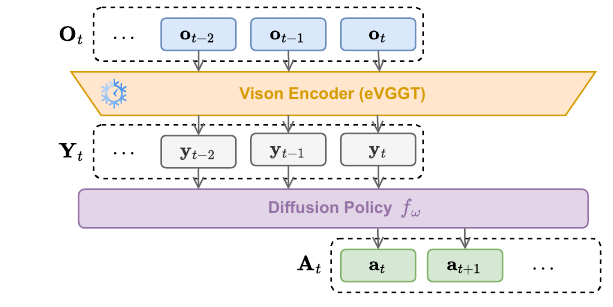}
    \caption{\textbf{Integration of eVGGT into DP.} eVGGT encodes observations to transfer geometric representation for the guided diffusion process.}
    \label{fig: dp-policy}
    \vspace{-1ex}
\end{figure}

We focus on two representative IL approaches: \textit{i)} ACT~\cite{zhao2023learning}, a transformer-based policy, and \textit{ii)} DP~\cite{chi2023diffusion}, a diffusion model-based policy. The integration of our eVGGT encoder into these policies is quite straightforward, as illustrated in Figs.~\ref{fig: act-policy} and~\ref{fig: dp-policy}.

\nbf{Action Chunking with Transformer (ACT).} To enforce temporal smoothness in predicted actions, ACT predicts the next chunk of $k$ actions, $\mathbf{a}_{t:t+k}$, from the current observation $\mathbf{o}_t$ (which may include multiple views) at timestep $t$. This is implemented using a transformer-based conditional VAE comprising two components: \textit{i)} a transformer encoder 
\(
f^{\text{enc}}_{\phi}(\mathbf{z}_t \mid \mathbf{o}_t, \mathbf{a}_{t:t+k})
\) 
and \textit{ii)} a transformer decoder 
\(
f^{\text{dec}}_{\varphi}(\hat{\mathbf{a}}_{t:t+k} \mid \mathbf{z}_t, \mathbf{o}_t)
\), 
where $\mathbf{z}_t$ is a latent variable.

The observation $\mathbf{o}_t$ is first embedded using eVGGT to produce a latent feature \(
\mathbf{y}_t = \text{eVGGT}(\mathbf{o}_t)
\) with geometric reasoning capabilities, which is then incorporated into both the encoder and decoder. Specifically, the transformer encoder maps the sequence 
\(
[\text{CLS}, \mathbf{y}_t, \mathbf{a}_{t:t+k}]
\) 
to the latent variable $\mathbf{z}_t$ extracted from the CLS token, while the transformer decoder takes the sequence 
\(
[\mathbf{z}_t, \mathbf{y}_t]
\) 
as input \textit{key} and generates the predicted action chunk $\hat{\mathbf{a}}_{t:t+k}$. During inference, the next action is computed as a weighted average over the predicted action chunk of length $k$.

\nbf{Diffusion Policy (DP).} 
DP formulates the policy as a denoising diffusion probabilistic model (DDPM)~\cite{ho2020denoising}, following the same principle as ACT by predicting a sequence of actions 
$\mathbf{A}_t = \mathbf{a}_{t:t+T_a}$ from a sequence of observations $\mathbf{O}_t = \mathbf{o}_{t-T_o:t}$, where $T_o$ and $T_a$ denote the respective chunk sizes as defined in~\cite{chi2023diffusion}. The process consists of two stages: a \textit{i)} forward process that gradually adds noise to the action sequence, and \textit{ii)} a reverse process that learns to recover the original action sequence, conditioning on the current observations.

In the forward process, the clean action chunk $\mathbf{A}_t$ is gradually perturbed with Gaussian noise according to a predefined schedule. At each diffusion step $k \in \{1, \dots, K\}$, the noisy data $\mathbf{A}_t^{(k)}$ is obtained as follows:
\begin{equation}
    \mathbf{A}_t^{(k)} = \sqrt{\overline{\alpha}_k}\,\mathbf{A}^{(0)}_t + \sqrt{1-\overline{\alpha}_k}\,\boldsymbol{\epsilon}, \quad \boldsymbol{\epsilon} \sim \mathcal{N}(\mathbf{0},\mathbf{I}),
\end{equation}
where $\overline{\alpha}_k=\prod_{j\leq k}{\alpha_j}$ with $\alpha_j$ defined by the variance schedule~\cite{ho2020denoising}, and $\mathbf{A}^{(0)}_t$ denotes as the original action chunk. 
In the reverse process, the observation $\mathbf{O}_t$ is first encoded into a global-context embedding $\mathbf{Y}_t = \text{eVGGT}(\mathbf{O}_t)$, which provides informative 3D guidance for denoising. Starting from noisy $\mathbf{A}_t^{(k)}$, we iteratively denoise toward $\mathbf{A}_t^{(0)}$ by:
\begin{equation}
    \mathbf{A}^{(k-1)}_t = \frac{1}{\sqrt{\alpha_k}} 
    \left(\mathbf{A}^{(k)}_t - \frac{1 - \alpha_k}{\sqrt{1 - \overline{\alpha}_k}} 
    f_{\omega}(\mathbf{A}^{(k)}_t, \mathbf{Y}_t, k)\right) 
    + \sigma_k\boldsymbol{z},
\end{equation}
where $\boldsymbol{z} \sim \mathcal{N}(\mathbf{0},\mathbf{I})$, $f_{\omega}$ denotes the transformer-based denoising network~\cite{chi2023diffusion}, and $\sigma_k$ represents the standard deviation of the Gaussian noise in the reverse process~\cite{ho2020denoising}.
\section{EXPERIMENT}
\label{sec: experiment}
We perform experiments to evaluate the effectiveness of our proposed efficient geometry-aware visuomotor policies. Specifically, we aim to answer the following questions:

\begin{itemize}[leftmargin=*]
    \item \textbf{Q1:} To what extent does our geometry-aware eVGGT encoder improve robotic manipulation performance? 
    \item \textbf{Q2:} How does the geometric prior in eVGGT compare to standard 2D and geometry-aware encoders in terms of task success and sample efficiency?
    \item \textbf{Q3:} Can our approach transfer effectively from simulation to the real world?
\end{itemize}

\begin{table*}[!ht]
    \centering
    \caption{Success rates on RoboTwin simulator}
    \label{tab: robotwin_policy_success_final}
    \vspace{1ex}
    \setlength{\tabcolsep}{3pt}
    \resizebox{\linewidth}{!}{
    \begin{tabular}{rc ccc ccc ccc ccc ccc ccc ccc}
        \toprule
        Method & Modality & \multicolumn{3}{c}{Beat Block Hammer} & \multicolumn{3}{c}{Adjust Bottle} & \multicolumn{3}{c}{Lift Pot} & \multicolumn{3}{c}{Place Burger Fries} & \multicolumn{3}{c}{Press Stapler} & \multicolumn{3}{c}{Handover Mic} & \multicolumn{3}{c}{\textbf{Average}} \\
        \cmidrule(lr){3-5} \cmidrule(lr){6-8} \cmidrule(lr){9-11} \cmidrule(lr){12-14} \cmidrule(lr){15-17} \cmidrule(lr){18-20} \cmidrule(lr){21-23}
        & & Seen & Unseen & Avg & Seen & Unseen & Avg & Seen & Unseen & Avg & Seen & Unseen & Avg & Seen & Unseen & Avg & Seen & Unseen & Avg & Seen & Unseen & Avg \\
        \midrule
        DP3~\cite{ze20243d}     & Point cloud & 0.082 & \underline{0.018} & 0.050 & \textbf{0.622} & 0.296 & \underline{0.459} & 0.102 & \underline{0.038} & 0.070 & 0.184 & \underline{0.036} & 0.110 & 0.152 & \underline{0.238} & \underline{0.195} & \textbf{0.724} & \textbf{0.316} & \textbf{0.520} & \textbf{0.311} & \underline{0.157} & \textbf{0.234} \\ 
        RDT~\cite{liurdt}     & RGB  & 0.090 & \textbf{0.092} & \underline{0.091} & 0.496 & \underline{0.354} & 0.425 & \underline{0.118} & 0.036 & \underline{0.077} & \underline{0.196} & \textbf{0.134} & \textbf{0.165} & \underline{0.198} & 0.122 & 0.160 & \underline{0.606} & 0.204 & \underline{0.405} & 0.284 & 0.157 & 0.221 \\ \midrule
        ACT~\cite{zhao2023learning}      & RGB & 0.074 & 0.000 & 0.037 & 0.162 & 0.118 & 0.140 & 0.034 & 0.006 & 0.020 & 0.042 & 0.000 & 0.021 & 0.024 & 0.016 & 0.020 & 0.022 & 0.000 & 0.011 & 0.060 & 0.023 & 0.042 \\
        ACT+eVGGT (ours) & RGB & 0.066 & 0.012 & 0.039 & 0.198 & 0.104 & 0.151 & 0.016 & 0.022 & 0.019 & 0.098 & 0.000 & 0.049 & 0.136 & 0.032 & 0.084 & 0.038 & 0.014 & 0.026 & 0.092 & 0.031 & 0.061 \\ \midrule
        DP~\cite{chi2023diffusion}      & RGB  & \underline{0.142} & 0.014 & 0.078 & 0.462 & 0.238 & 0.350 & 0.062 & 0.020 & 0.041 & 0.164 & 0.038 & 0.101 & 0.182 & 0.094 & 0.138 & 0.502 & 0.118 & 0.310 & 0.252 & 0.087 & 0.170 \\
        DP+eVGGT (ours)    & RGB  & \textbf{0.178} & 0.012 & \textbf{0.095} & \underline{0.528} & \textbf{0.404} & \textbf{0.466} & \textbf{0.138} & \textbf{0.040} & \textbf{0.089} & \textbf{0.256} & 0.012 & \underline{0.134} & \textbf{0.202} & \textbf{0.262} & \textbf{0.232} & 0.518 & \underline{0.244} & 0.381 & \underline{0.303} & \textbf{0.162} & \underline{0.233} \\
        \bottomrule
    \end{tabular}
    }
\end{table*}

\subsection{Experimental Setup}\label{sec: robot-manipulation-exp}
This section details the KD training of eVGGT and the protocol for evaluating downstream robotic manipulation success rates.

\nbf{Training Data.} For the upstream distillation of eVGGT from VGGT, we use 40 training tasks in RoboTwin~\cite{chen2025robotwin} (10 demonstrations/task, 4 camera views, $\sim$320K total images) and 3 tasks in ManiSkill~\cite{taomaniskill3} (\textit{Stack Pyramid}, \textit{Peg Insertion Side}, and \textit{Push T}; 100 demonstrations/task, single view, $\sim$60K total images). This process is highly scalable as it relies strictly on label-free RGB inputs without depth or 3D ground truth. Distillation training was completed on four NVIDIA A100 GPUs (40GB) over six days. For downstream IL evaluation, we select 6 representative RoboTwin tasks (\textit{Beat Block Hammer, Adjust Bottle, Lift Pot, Place Burger Fries, Press Stapler, Handover Mic}) and 3 ManiSkill tasks (\textit{Pick Cube, Push Cube, Stack Cube}). 

\nbf{Evaluation Protocol.} In RoboTwin, the original benchmark trains on \textit{clean} demonstrations and evaluates on \textit{randomized} (cluttered) scenes. We argue that cluttered scenes provide richer information regarding object dynamics~\cite{zhu2023viola}. Thus, we modify the protocol: policies are trained on 50 \textit{randomized} demonstrations for 10K iterations and evaluated in both seen (randomized) and unseen (clean) settings. For ManiSkill, we follow standard setups~\cite{liurdt, vuong2025action} and train for 30K iterations. Following common evaluation practice, we run 100 trials per RoboTwin task~\cite{chen2025robotwin} and 50 trials per ManiSkill task~\cite{taomaniskill3}, across 5 different seeds.

\nbf{Baselines.} We evaluate our geometry-integrated policies, ACT+eVGGT and DP+eVGGT, against several state-of-the-art imitation learning frameworks. These include RDT~\cite{liurdt}, a high-performance baseline comparable to modern VLA models, as well as the standard ACT~\cite{zhao2023learning} and Diffusion Policy (DP)~\cite{chi2023diffusion}. For the RoboTwin benchmark, we also include DP3~\cite{ze20243d}, though it is excluded from ManiSkill due to the lack of native 3D point cloud inputs in that setup~\cite{taomaniskill3}.

To evaluate our geometry-aware encoder, we compare DP+eVGGT on RoboTwin against other geometry-aware visuomotor policies, including DP+MoGe~\cite{wang2025moge}, DP+Metric3Dv2~\cite{hu2024metric3d}, and the full DP+VGGT~\cite{wang2025vggt}. All models are trained on a single NVIDIA RTX 4090 GPU (24GB). Notably, while most methods converge in under 12 hours per task, the full DP+VGGT requires approximately 7 days per task due to the massive computational overhead of the original VGGT encoder. This highlights the practical efficiency of our proposed eVGGT adaptation.

\subsection{Robotic Manipulation in Simulation}\label{sec: robot-manipulation-exp}

\begin{table}[!ht]
    \centering
    \caption{Success rates on ManiSkill Simulator}
    \label{tab: maniskill-success}
    \setlength{\tabcolsep}{5.6pt} 
    \vspace{1ex}
    \resizebox{\linewidth}{!}{
    \begin{tabular}{rccccc}
        \toprule
        Method   & Modality & Pick Cube & Push Cube & Stack Cube & \textbf{Average} \\
        \midrule
        DP~\cite{chi2023diffusion} & RGBD & \underline{0.952} & \underline{0.922} & 0.524 & 0.799 \\ \midrule
        RDT~\cite{liurdt}      & RGB & 0.774 & \textbf{1.000} & \textbf{0.702} & \underline{0.825} \\ \midrule
        ACT~\cite{zhao2023learning}      & RGB & 0.712 & 0.904 & 0.474 & 0.697 \\
        ACT+eVGGT (ours) & RGB & 0.756 & 0.888 & 0.496 & 0.713 \\ \midrule
        DP~\cite{chi2023diffusion}       & RGB & 0.848 & 0.876 & 0.618 & 0.781 \\
        DP+eVGGT (ours)  & RGB & \textbf{0.964} & 0.912 & \underline{0.654} & \textbf{0.843} \\
        \bottomrule
    \end{tabular}
    }
\end{table}

\nbf{Results on RoboTwin.} On RoboTwin, integrating eVGGT into existing IL frameworks yields substantial performance gains, as summarized in Table~\ref{tab: robotwin_policy_success_final} (\textbf{bold} indicates the best result, and \underline{underline} indicates the second-best). Specifically, ACT+eVGGT and DP+eVGGT achieve improvements of 1.9\% and 6.3\% over the original ACT and DP, respectively, highlighting the effectiveness of the encoded 3D knowledge in robotic policy learning. Notably, DP+eVGGT surpasses RDT by 1.2\%, indicating that geometric representations can significantly improve non-pretrained policies, even outperforming large-scale pretrained methods using conventional vision encoders. While DP3 achieves the highest performance (23.4\%) by using clean, post-processed point clouds that remove walls and desks, the success rate of DP+eVGGT reaches a comparable 23.3\% using only RGB inputs. This suggests that RGB-only policies with a strong geometry-aware encoder can approach the performance of point cloud-based methods without requiring post-processed data.

\nbf{Results on ManiSkill.} Table~\ref{tab: maniskill-success} presents the effects of incorporating eVGGT into existing IL frameworks on ManiSkill. Notably, DP+eVGGT achieves the highest performance, outperforming RDT and DP with RGBD input by 1.8\% and 4.4\%, respectively. This result indicates that 3D spatial representations can support policy success rates at a level comparable to the pretrained knowledge utilized by RDT and the additional depth modality employed by DP. Similar to the results on RoboTwin, ACT+eVGGT and DP+eVGGT achieve average improvements of 1.6\% and 6.2\% over the original ACT and DP, respectively. Overall, our eVGGT encoder substantially strengthens performance across diverse manipulation settings.

\begin{table*}[!ht]
    \centering
    \caption{Comparison with Different Vision Methods}
    \label{tab: vision_encoder_success_final_best}
    \vspace{1ex}
    \resizebox{\linewidth}{!}{
    \setlength{\tabcolsep}{4pt}
    \begin{tabular}{c|l|ccc|ccc}
        \toprule
        \multicolumn{2}{c|}{} & \multicolumn{3}{c}{\textbf{Modality \& Efficiency}} & \multicolumn{3}{c}{\textbf{Success Rate}} \\
        \cmidrule(lr){3-5} \cmidrule(lr){6-8}
        \multicolumn{2}{c|}{Method} & Modality & Inference Speed (Hz) & Memory (GB) & Beat Block Hammer & Adjust Bottle & \textbf{Average} \\
        \midrule
        \parbox[t]{2mm}{\multirow{2}{*}{\rotatebox[origin=c]{90}{\scriptsize 3D}}} 
        & DP3~\cite{ze20243d}              & Point cloud (simulator) & - & 2.54 & 0.082 & \textbf{0.622} & \underline{0.352} \\
        & DP3~\cite{ze20243d}              & Point cloud (VGGT~\cite{wang2025vggt}) & 3.5 & 10.47 & 0.042 & 0.252 & 0.147 \\ \midrule
        
        \parbox[t]{2mm}{\multirow{2}{*}{\rotatebox[origin=c]{90}{\scriptsize 2D}}} 
        & DP~\cite{chi2023diffusion}+ResNet & RGB       & 25.4 & 2.00 & 0.142 & 0.462 & 0.302 \\
        & DP~\cite{chi2023diffusion}+ViT    & RGB       & 24.8 & 2.11 & 0.124 & 0.374 & 0.249 \\ \midrule
        
        \parbox[t]{2mm}{\multirow{4}{*}{\rotatebox[origin=c]{90}{\scriptsize \shortstack{geo-aware}}}} 
        & DP+Metric3Dv2~\cite{hu2024metric3d} & RGB       & 17.1 & 4.39 & 0.138 & 0.392 & 0.265 \\
        & DP+MoGe~\cite{wang2025moge}      & RGB       & 17.7 & 5.63 & 0.152 & 0.412 & 0.282 \\
        & DP+VGGT~\cite{wang2025vggt}      & RGB       & 4.5 & 8.14 & \underline{0.176} & 0.512 & 0.344 \\
        & DP+eVGGT (ours)                  & RGB       & 20.9 & 3.01 & \textbf{0.178} & \underline{0.528} & \textbf{0.353} \\
        \bottomrule
    \end{tabular}
    }
\end{table*}

\nbf{Comparison of Vision Methods.} To isolate the impact of our proposed geometry-aware vision encoder, we compare against three representative categories: \textit{i)} 3D point cloud (DP3), \textit{ii)} standard 2D RGB (ResNet, ViT), and \textit{iii)} geometry-aware RGB (Metric3Dv2, MoGe, VGGT). Due to the excessively computational overhead of DP+VGGT, we restrict its evaluation to two representative RoboTwin tasks: \textit{Beat Block Hammer} and \textit{Adjust Bottle}. The results in Table~\ref{tab: vision_encoder_success_final_best} reveal two key performance observations. First, DP3’s performance drops by 20.5\% when using reconstructed point clouds from VGGT instead of simulator-provided point clouds, as it relies on clean, post-processed point clouds rather than noisy, unstructured reconstructed point clouds. This highlights that our implicit encoding design is more robust. Second, across both standard RGB and geometry-aware RGB baselines, DP+eVGGT consistently improves the success rate, exceeding DP+ResNet by 5.1\% and DP+ViT by 10.4\%, and achieving the best performance among geometry-aware encoders (35.3\%). These results suggest that encoding geometry-aware scene dynamics is critical for manipulation, and that our encoder effectively preserves the VGGT-derived geometric priors needed for robot control.

\nbf{Computational Efficiency.} Table~\ref{tab: vision_encoder_success_final_best} also presents the computational overhead across all methods, measured using an NVIDIA RTX 4090 GPU within the RoboTwin simulator environment. The results indicate that DP+eVGGT achieves a frequency of 20.9\,Hz, closely matching the 25.4\,Hz of standard DP widely used in the literature~\cite{chi2023diffusion, wang2025one}. In contrast, all other geometry-aware baselines suffer from significant latency and memory overhead. DP+eVGGT is at least 18\% faster than the next fastest geometry-aware encoder (MoGe) and substantially $\approx$5$\times$ faster than the original VGGT. Furthermore, DP+eVGGT saves 1.3–5.1~GB compared to other geometry-aware methods. Critically, this reduced memory footprint is essential for real-world robotics, where the vision encoder must share limited GPU resources with concurrent processes such as motion planners and low-level controllers. The result confirms that eVGGT effectively resolves the latency bottleneck of geometric encoders for real-world deployment.

\subsection{Ablation Study}

To understand how our proposed components contribute to the visuomotor policy performance, we conduct an ablation study focusing on multi-view scalability and the impact of depth reconstruction accuracy on downstream policy success.

\begin{figure}[!ht]
    \centering
    \includegraphics[width=\linewidth]{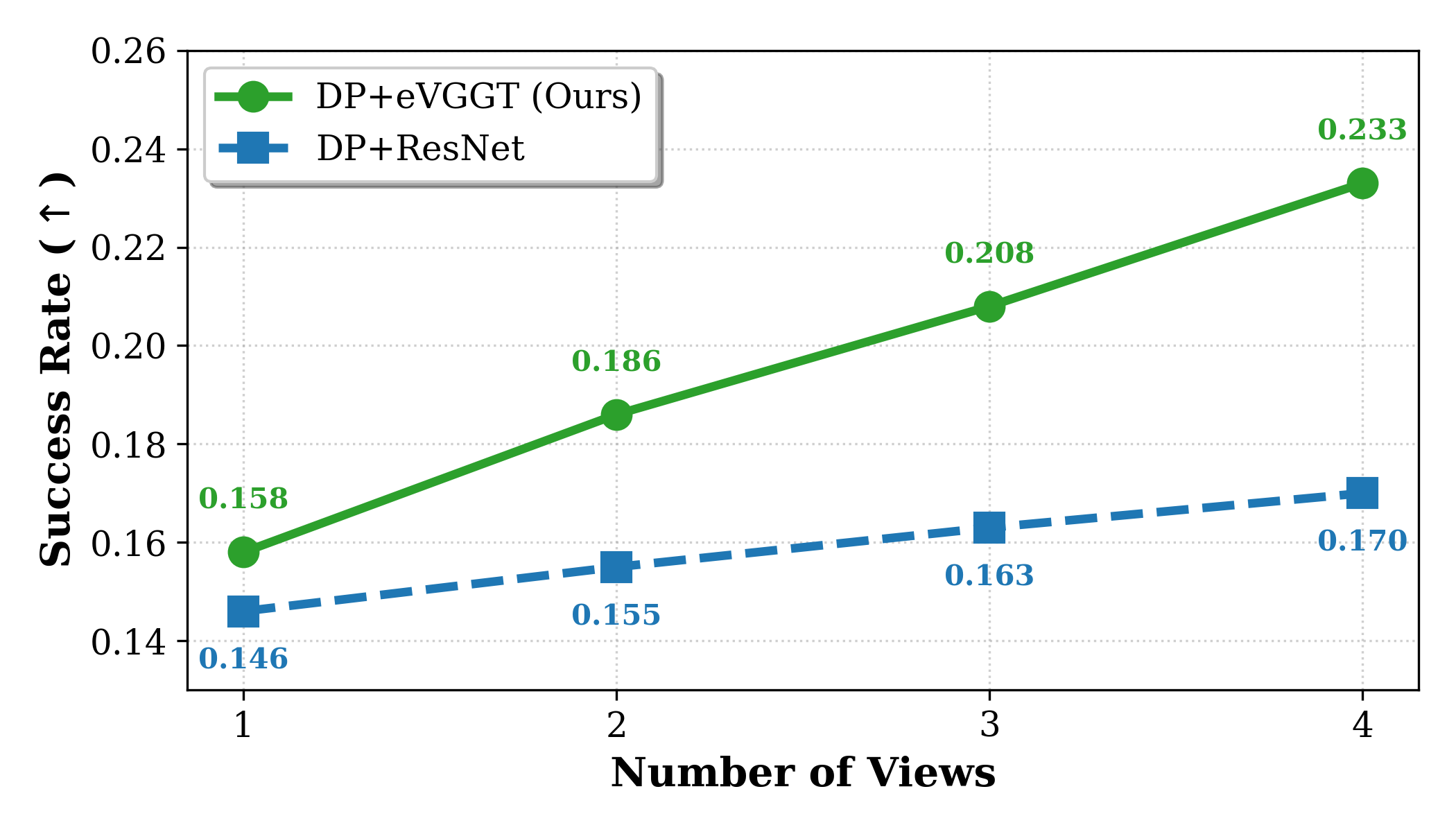}
    \caption{\textbf{Success rates when scaling views.} We compare DP+eVGGT against the DP+ResNet baseline across 1 to 4 RGB views on RoboTwin.}
    \label{fig: scaling-multi-view}
\end{figure}

\nbf{Multi-View Scaling.}
We first investigate the scalability of our proposed eVGGT encoder when provided with multiple RGB views, comparing it against a standard ResNet baseline. Fig.~\ref{fig: scaling-multi-view} shows the success rates on the RoboTwin as the number of input views scales from 1 to 4. While both architectures benefit from additional views, DP+eVGGT demonstrates significantly better scaling properties. Specifically, DP+eVGGT demonstrates significantly better scalability, achieving a 47.5\% improvement in success rate when scaling from one to four views. In contrast, the DP+ResNet baseline shows a smaller gain, achieving only 16.4\% improvement under the same conditions. At four views, our DP+eVGGT outperforms the baseline by 37.1\%, highlighting the advantage of geometry-aware features for complex multi-view inputs over standard latent visual encoders.
\begin{figure}
    \centering
    \includegraphics[width=\linewidth]{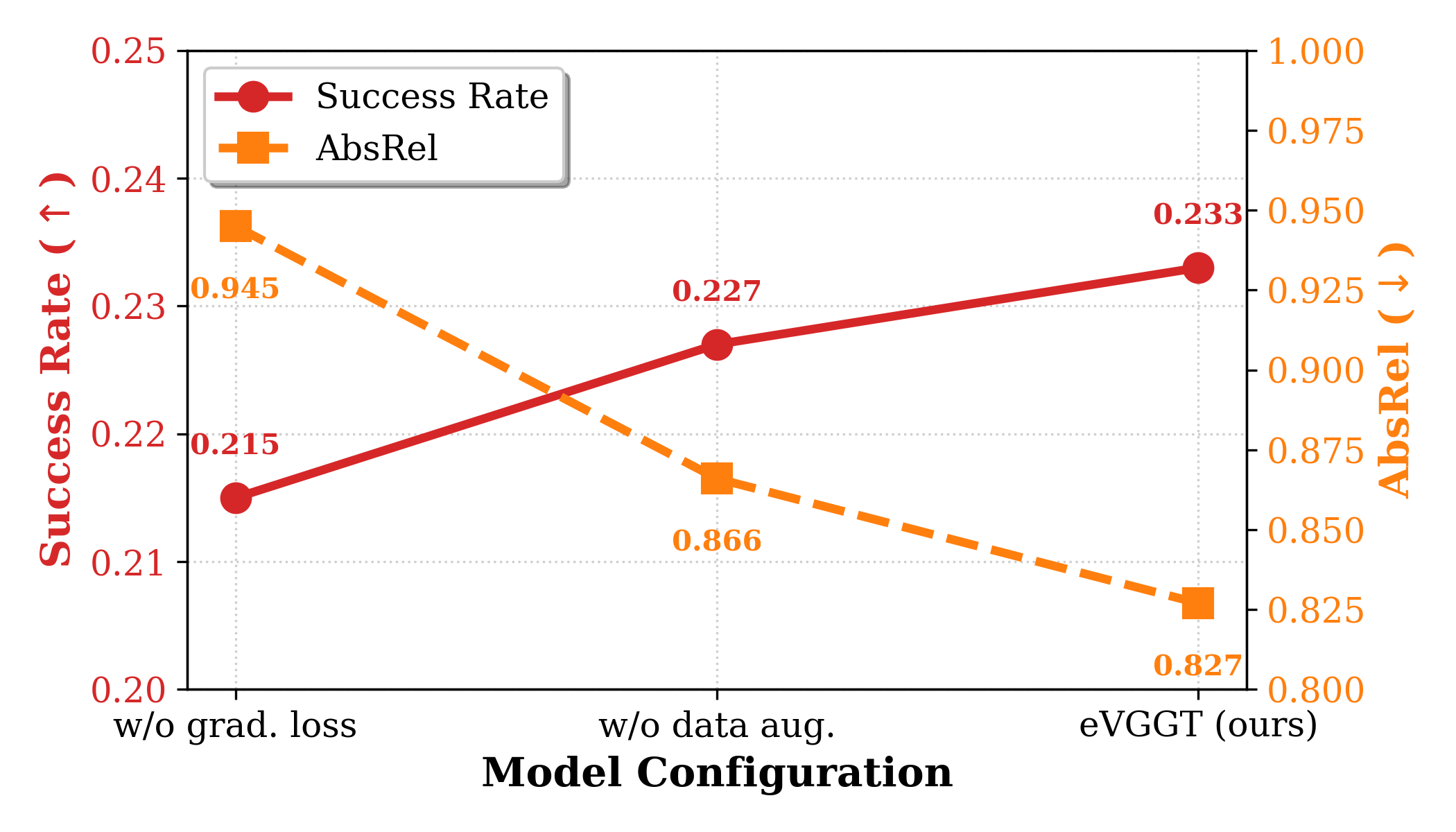}
    \caption{\textbf{Ablation study: Reconstruction accuracy vs. Success rates.} We compare AbsRel across three model configurations and their corresponding success rates when paired with DP.}
    \label{fig:ablation_correlation}
\end{figure}

\nbf{Impact of Proposed KD Framework.} We assess the impact of gradient loss (Eq.~\ref{eq: grad_loss}) and data augmentation (Sec.~\ref{sec: method-eVGGT}) on geometric accuracy (AbsRel~\cite{wang2025vggt}) and manipulation performance. As shown in Fig.~\ref{fig:ablation_correlation}, task success is strongly correlated with depth reconstruction quality: removing either component increases the AbsRel (to 0.945 and 0.866, respectively) and degrades success rates. Our full eVGGT model achieves the best reconstruction (0.827 AbsRel) and highest success rate (23.3\%), validating that precise, geometry-aware representations are pivotal for robust manipulation.

\begin{figure}[!ht]
    \centering

    \begin{minipage}{\linewidth}
        \centering
        \includegraphics[width=\linewidth, trim={0 0pt 0 10pt}, clip]{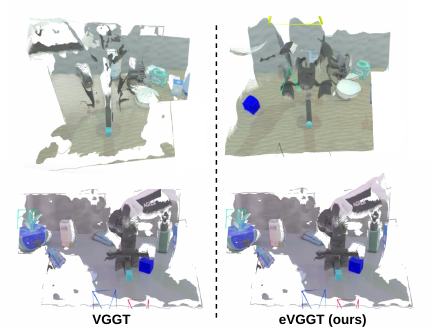}
        {\footnotesize (a) Point cloud visualization}
        \label{fig:point_cloud_vis}
    \end{minipage}

    \vspace{1em} 

    \begin{minipage}{\linewidth}
        \centering
        \includegraphics[width=\linewidth]{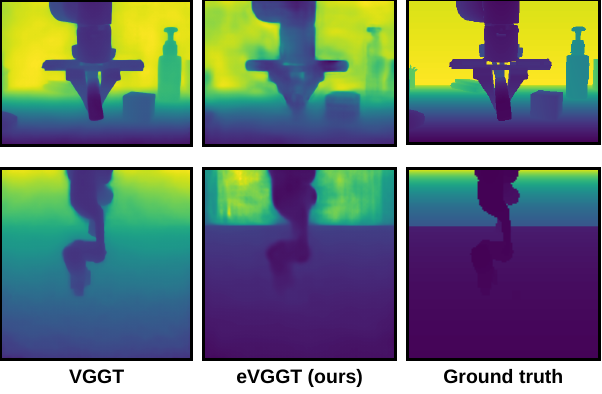}
        {\footnotesize (b) Depth prediction}
        \label{fig:depth_vis}
    \end{minipage}

    \vspace{1em}
     \caption{\textbf{Qualitative 3D reconstruction results.} Qualitative comparison between eVGGT and its teacher VGGT on 3D reconstruction. For point clouds, we omit the ground truth since it is provided as semantically labeled points rather than a comparable geometric reference. For depth, results are shown on RoboTwin (top) and ManiSkill (bottom).}
    \label{fig:qualitative_3d_recon}
\end{figure}

\nbf{3D Reconstruction Evaluation.} We qualitatively validate the 3D reconstruction performance of the proposed eVGGT relative to its teacher, VGGT, which achieves strong reconstruction accuracy and provides a robust backbone for learning geometric representations~\cite{keetha2025mapanything}. As shown in Fig.~\ref{fig:qualitative_3d_recon}, eVGGT matches the teacher’s reconstruction quality in both point-cloud generation and depth prediction. Notably, eVGGT maintains high geometric fidelity for structures critical to reliable robotic execution, accurately recovering the gripper arm and the interacted objects, including cubes, pots, and hammers, in both the reconstructed point clouds and the predicted depth maps. These results suggest that, despite its significantly reduced inference latency (Table~\ref{tab: vision_encoder_success_final_best}), eVGGT preserves the teacher's spatial reasoning capabilities, providing effective geometric priors for downstream visuomotor policies. Additional results are included in the Supplementary Video.

\begin{figure}[!ht]
\centering
\scalebox{0.9}{
\def\svgwidth{1\columnwidth}
\begingroup%
  \makeatletter%
  \providecommand\color[2][]{%
    \errmessage{(Inkscape) Color is used for the text in Inkscape, but the package 'color.sty' is not loaded}%
    \renewcommand\color[2][]{}%
  }%
  \providecommand\transparent[1]{%
    \errmessage{(Inkscape) Transparency is used (non-zero) for the text in Inkscape, but the package 'transparent.sty' is not loaded}%
    \renewcommand\transparent[1]{}%
  }%
  \providecommand\rotatebox[2]{#2}%
  \newcommand*\fsize{\dimexpr\f@size pt\relax}%
  \newcommand*\lineheight[1]{\fontsize{\fsize}{#1\fsize}\selectfont}%
  \ifx\svgwidth\undefined%
    \setlength{\unitlength}{260.90922438bp}%
    \ifx\svgscale\undefined%
      \relax%
    \else%
      \setlength{\unitlength}{\unitlength * \real{\svgscale}}%
    \fi%
  \else%
    \setlength{\unitlength}{\svgwidth}%
  \fi%
  \global\let\svgwidth\undefined%
  \global\let\svgscale\undefined%
  \makeatother%
  \begin{picture}(1,0.75000002)%
    \lineheight{1}%
    \setlength\tabcolsep{0pt}%
    \put(0,0){\includegraphics[width=\unitlength,page=1]{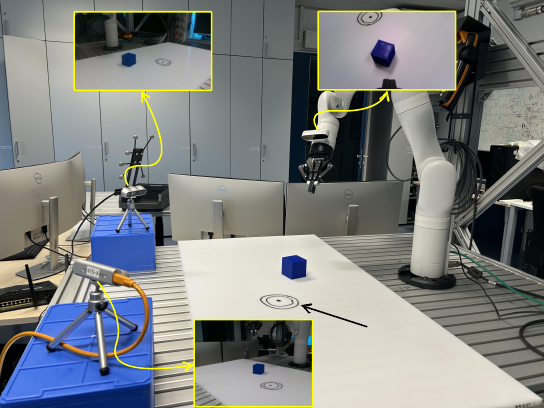}}%
    \put(0.68163768,0.13087716){\color[rgb]{0,0,0}\makebox(0,0)[lt]{\lineheight{1.25}\smash{\begin{tabular}[t]{l}target\end{tabular}}}}%
  \end{picture}%
\endgroup%

}
\vspace{1ex}
\caption{\textbf{Robot Setup.} We conduct robot experiments with three camera views to push the cube to the target location.}
\label{fig:robot_experiment}
\end{figure}

\begin{table}[!ht]
    \centering
    \caption{Real-world robot success rate}\label{tab:real_world_success}
    \vspace{1ex}
    \resizebox{\linewidth}{!}{
    \begin{tabular}{lccc}
    \toprule
    Method & \makecell{Push Cube} & \makecell{Pick \& Place Cube} & \makecell{Open Drawer} \\
    \midrule
    DP~\cite{chi2023diffusion}  & 0.54 & 0.40 & 0.42 \\
    DP+eVGGT (ours)            & \textbf{0.72} & \textbf{0.60} & \textbf{0.66} \\
    \bottomrule
    \end{tabular}
    }
\end{table}

\subsection{Real-World Robotic Experiments}
\nbf{Setup.} For the real-robot evaluation, we focus on three single-arm manipulation tasks using a Kinova Gen3 robot: \textit{i) Push Cube}, \textit{ii) Pick and Place Cube}, and \textit{iii)} \textit{Open Drawer}. Example robot setup is shown in Fig. \ref{fig:robot_experiment}. We collect $10$ human demonstrations per task via teleoperation using a 3Dconnexion SpaceMouse, allowing an operator to guide the robot end-effector in real time. To increase data diversity, we leverage MimicGen \cite{mandlekar2023mimicgen} to generate $\approx$ $2000$ additional synthetic demonstrations for each task in a simulated RoboCasa~\cite{nasiriany2024robocasa} environment. This yields a large-scale training dataset featuring diverse scene configurations for the Kinova Gen3. In the dataset generation, we record three camera viewpoints during each demonstration: an eye-in-hand RGB camera on the robot’s wrist and two static third-person RGB cameras providing external views.

\nbf{Baselines.} We train two policies: DP+eVGGT, using our geometry-aware encoder, and a DP~\cite{chi2023diffusion} baseline with a conventional ResNet encoder.
Both policies are trained on the same collected dataset. We evaluate 50 autonomous trials per task on hardware. A trial is successful if the robot pushes the cube past a designated target line (\textit{Push Cube}), places it within a 2 cm radius of the goal region (\textit{Pick and Place Cube}), or pulls the drawer until the interior compartment is fully visible from the primary camera view (\textit{Open Drawer}).


\nbf{Results.} Table~\ref{tab:real_world_success} reports the real-world success rates. Specifically, we observe absolute performance improvements of 18\% in \textit{Push Cube}, 20\% in \textit{Pick and Place Cube}, and 24\% in \textit{Open Drawer} compared to the baseline. These consistent gains indicate that the geometry-aware encoder (eVGGT) provides a superior 3D understanding that translates into more robust real-world manipulation. We provide a video of the real-robot experiments in the Supplementary Video.


\subsection{Remarks}
Tables~\ref{tab: robotwin_policy_success_final},~\ref{tab: maniskill-success}, and~\ref{tab: vision_encoder_success_final_best} show that using eVGGT as the vision encoder substantially improves manipulation success, achieving performance comparable to large-scale pretrained methods such as RDT~\cite{liurdt} and approaches utilizing additional 3D information like DP3~\cite{ze20243d} (\textbf{Q1}). Our eVGGT provides two key benefits: \textit{i) Scalable geometric priors:} Compared to 2D encoders, eVGGT scales more effectively with multi-view inputs (Fig.~\ref{fig: scaling-multi-view}). Furthermore, our proposed KD scheme significantly improves 3D reconstruction accuracy, leading to substantial gains in success rates (Fig.~\ref{fig:ablation_correlation}). \textit{ii) Inference efficiency:} eVGGT achieves a $5\times$ speedup over VGGT with significantly lower VRAM, while maintaining high performance (Table~\ref{tab: vision_encoder_success_final_best}, Fig.~\ref{fig:qualitative_3d_recon}), enabling practical for real-world deployment (\textbf{Q2}). Finally, real-robot evaluations (Table~\ref{tab:real_world_success}) show that the policy with our geometry-aware encoder successfully transfers from simulation to the real world, improving manipulation success rates (\textbf{Q3}).
\section{DISCUSSION}
We introduce eVGGT, a lightweight, geometry-aware vision encoder for visuomotor policies that significantly improves manipulation success over conventional 2D vision encoders. eVGGT delivers real-time inference with a reduced memory footprint while preserving strong spatial reasoning, making high-quality geometric priors practical for deployment on robots. Across complex manipulation tasks, our results show that eVGGT provides a robust and scalable backbone for visuomotor learning, offering an efficient alternative to standard vision encoders without sacrificing geometric fidelity.

\nbf{Limitation.} Although our proposed geometry-aware visuomotor policies show substantial improvements, the approach has notable limitations. First, our 3D representations are implicit. While this allows the model to capture dense geometric priors that outperform explicit point-cloud-based methods (as shown in Table~\ref{tab: vision_encoder_success_final_best}), it limits direct human interpretability of the latent space. We view this as a trade-off where performance gains currently outweigh the need for explicit 3D interpretability. Second, the explicit output of eVGGT is dense, unstructured point clouds, which may pose challenges for IL baselines that expect structured geometric input. Bridging the gap between our implicit features and explicit 3D reconstructions for downstream manipulation tasks remains a promising future direction.



\bibliographystyle{class/IEEEtran}
\bibliography{class/IEEEabrv,reference}

\end{document}